%% file: main.tex
\documentclass[lettersize,journal]{IEEEtran}
\usepackage{amsmath,amsfonts}
\usepackage{algorithmic}
\usepackage{algorithm}
\usepackage{array}
\usepackage[caption=false,font=normalsize,labelfont=sf,textfont=sf]{subfig}
\usepackage{textcomp}
\usepackage{stfloats}
\usepackage{url}
\usepackage{verbatim}
\usepackage{graphicx}
\usepackage{cite}
\usepackage{comment}
\usepackage{multirow}
\usepackage{multicol}
\usepackage{booktabs}
\usepackage{bbm}
\usepackage{color}
\usepackage{makecell}
\usepackage{comment}
\begin{document}

\newcommand{\pei}[1]{\textcolor[rgb]{0,0,1} {#1}}

\title{Activating the Discriminability of Novel Classes for Few-shot Segmentation} 

\author{Dianwen Mei\textsuperscript{\dag}, 
        Wei Zhuo\textsuperscript{\dag}, 
        Jiandong Tian,
        Guangming Lu,~\IEEEmembership{Member,~IEEE}
        and
        Wenjie Pei$^*$
\thanks{\textsuperscript{\dag} Equal contribution.}
\thanks{$^*$ Wenjie Pei is the corresponding author.}
\thanks{Dianwen Mei, Guangming Lu and Wenjie Pei are with the Department of
Computer Science, Harbin Institute of Technology at Shenzhen, Shenzhen
518057, China (e-mail: 178mdw@gmail.com; luguangm@hit.edu.cn; wenjiecoder@outlook.com).}%
\thanks{Wei zhuo is with Tencent, China (e-mail: wei.zhuowx@gmail.com).}
\thanks{Jiandong Tian is with Shenyang Institute of Automation, Chinese Academy of Sciences (e-mail: tianjd@sia.cn).}
}




\maketitle

\input{Abstract}

\begin{IEEEkeywords}
Few-shot segmentation, prototype matching, pseudo labeling, activating the discriminability.
\end{IEEEkeywords}

\section{Introduction}
\input{Introduction}

\section{Related Work}
\input{Related_work}

\section{OUR APPROACH}
\input{Method}

\section{Experiments}
\input{Experiment}

\section{Conclusion}
\input{Conclusion}

\bibliographystyle{IEEEtran}
\bibliography{ref.bib}

\vfill

\end{document}

%% file: Abstract.tex
\begin{abstract}

Despite the remarkable success of existing methods for few-shot segmentation, there remain two crucial challenges. First, the feature learning for novel classes is suppressed during the training on base classes in that the novel classes are always treated as background. Thus, the semantics of novel classes are not well learned. Second, most of existing methods fail to consider the underlying semantic gap between the support and the query resulting from the representative bias by the scarce support samples. To circumvent these two challenges, we propose to activate the discriminability of novel classes explicitly in both the feature encoding stage and the prediction stage for segmentation. In the feature encoding stage, we design the Semantic-Preserving Feature Learning module (\emph{SPFL}) to first exploit and then retain the latent semantics contained in the whole input image, especially those in the background that belong to novel classes. In the prediction stage for segmentation, we learn an Self-Refined Online Foreground-Background classifier (\emph{SROFB}), which is able to refine itself using the high-confidence pixels of query image to facilitate its adaptation to the query image and bridge the support-query semantic gap. Extensive experiments on PASCAL-5$^i$ and COCO-20$^i$ datasets demonstrates the advantages of these two novel designs both quantitatively and qualitatively.

\end{abstract}

%% file: Introduction.tex
\IEEEPARstart{W}{hile} semantic segmentation based on deep learning has achieved remarkable progress~\cite{long2015fully, yuan2020object, chen2018encoder, zhao2017pyramid}, it entails a large amount of mask-annotated data for supervised learning which is extremely exhaustive and expensive. Few-shot semantic segmentation is posed to address this problem by adapting a pre-trained segmentation model on base classes to novel classes using only a few annotated samples, namely support samples.

Most of existing methods for few-shot segmentation follow the prototype-matching paradigm~\cite{siam2019adaptive,wang2019panet, yang2021mining,liu2020part, lu2021simpler}, which performs support-query metric learning and conducts segmentation by measuring the semantic similarities between each pixel of the query image and the foreground and background prototypes learned from the support images. This type of methods either focus on learning representative prototypes from support images~\cite{wang2019panet, liu2020part}, or seek to design effective semantic matching mechanism~\cite{yang2021mining, lu2021simpler}. In contrast to such prototype-matching paradigm, another typical way of few-shot semantic segmentation~\cite{zhang2019canet, yang2020prototype, liu2020crnet, tian2020prior, li2020fss}, which we call the parametric relation-decoding paradigm, is to capture the semantic relation between the support and query by constructing a parametric decoder.

\begin{figure}[!t]
\centering
\includegraphics[width=1.0\linewidth]{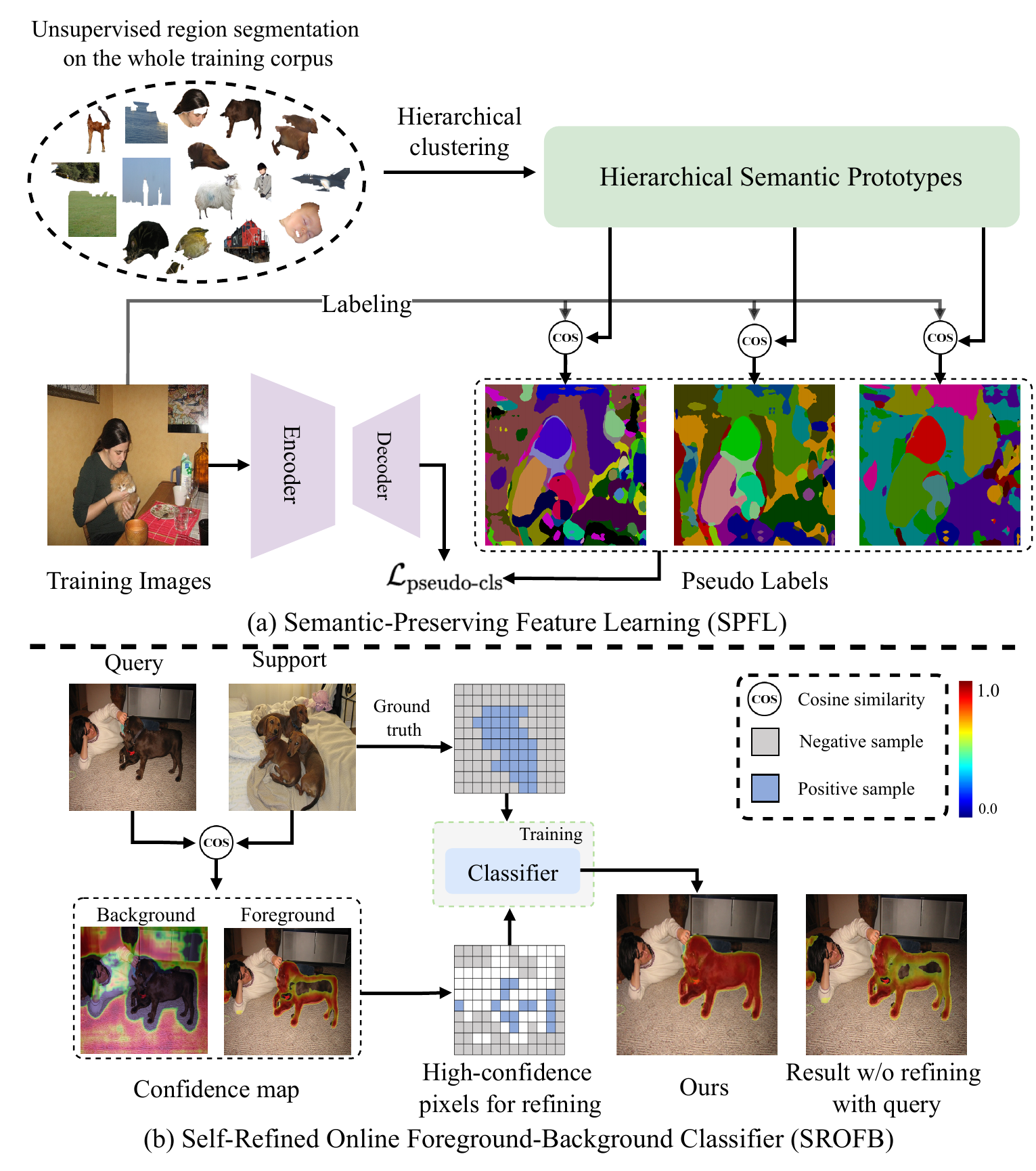}
\caption{Our method consists of two core modules, namely \emph{SPFL} and \emph{SROFB}, for activating the discriminability of novel classes in the feature encoding stage and the prediction stage for segmentation, respectively. \emph{SPFL} first performs unsupervised region segmentation and extracts hierarchical semantic prototypes from the entire training corpus to exploit latent semantics, which are leveraged to conduct pixel-level pseudo labeling for an input image. Then \emph{SPFL} conducts supervised pixel-level classification to learn an effective feature encoder that can preserve the latent semantics in the image. In the prediction stage, \emph{SROFB} performs pixel-wise binary classification for segmentation. It is not only trained based on the support images, but also refined using the high-confidence pixels of the query image. As a result, \emph{SROFB} can adapt to the query image smoothly and thus bridge the potential support-query semantic gap.
}
\label{fig:teaser}
\end{figure}

Despite the notable success of aforementioned two types of methods, there remain two potential limitations. The first limitation lies in the insufficient feature learning for novel classes, arising from the common implementation of these methods that all regions other than the foreground are considered as background during the training on base classes. Thus, the objects of novel classes are also treated as background, which suppresses the feature learning capability of these methods for novel classes and adversely affects the segmentation performance on novel classes. Second, these methods attempt to segment novel classes by parsing either the similarity or the semantic relation between the support and the query, failing to consider the underlying semantic gap between them resulting from the representative bias by the scarce support samples. In this work we propose two novel designs to deal with these two limitations. 



To enhance the feature learning capability on novel classes, we propose to activate the discriminability of novel classes explicitly in both the feature encoding stage and the prediction stage for segmentation, as shown in Figure~\ref{fig:teaser}. In the feature encoding stage, we design the Semantic-Preserving Feature Learning module (\emph{SPFL}) to first exploit and then retain latent semantics contained in the whole input image, especially those in the background that may belong to novel classes during the training on base classes. Specifically, the proposed \emph{SPFL} module first performs unsupervised region segmentation and extracts semantic prototypes from the entire training data to explore latent semantics. Then the obtained semantic prototypes are leveraged to conduct pixel-level pseudo labeling for each image by assigning each pixel to the semantically nearest prototype. Finally, \emph{SPFL} performs supervised pixel-level classification using the pseudo labels, which can be considered as a pretext task, to learn an effective encoder that can preserve the explored latent semantics including those from both the base and novel classes. In particular, to exploit the multi-scale latent semantics, our \emph{SPFL} produces hierarchical semantic prototypes and then perform multi-granularity pseudo labeling for supervised learning. Such strategy of exploring latent semantics in the background has been studied previously in MLC~\cite{yang2021mining}. However, MLC only aims to mine those latent semantics that are similar to foreground semantics of base classes, which substantially differs from our method.

To activate the discriminability of novel class during the prediction stage for segmentation, we learn the Self-Refined Online Foreground-Background classifier (\emph{SROFB}) to perform pixel-wise binary classification for segmentation. It is different from the strategy of modeling the support-query similarity metric or relation, adopted in the prototype-matching paradigm or the relation-decoding paradigm. Since the \emph{SROFB} classifier is learned online for each novel class individually, its performance is not limited by the base-to-novel generalization performance, which is typically suffered by the prototype-matching or relation-decoding paradigms.

A straightforward way of training the \emph{SROFB} classifier is to use the labeled pixel data in the support images. Nevertheless, training on the scarce support images is prone to overfitting on these samples. Moreover, the semantic gap between the support images and the query image, the second limitation of the prototype-matching or relation-decoding paradigms described above, 
limits the generalization performance of the classifier on the query image. To tackle these downsides, we propose to refine the \emph{SROFB} classifier using (the pixels of) the query image besides the training on support images. Since the segmentation annotation of the query images are not provided, we first perform a rough segmentation based on the support-query matching following the prototype-matching paradigm. Then we select both the foreground and background pixels with high matching score in the query images as the positive and negative training samples to refine \emph{SROFB}, respectively. As a result, the proposed \emph{SROFB} classifier is able to adapt to the query image smoothly for segmentation.

To summarize, we make following contributions:
\begin{itemize}
\item We propose the Semantic-Preserving Feature Learning (\emph{SPFL}) module to exploit and retain the latent semantics, especially those of novel classes contained in the background during the training on base classes. Thus, our model can learn an effective feature encoder which is able to extract discriminative features for novel classes.
\item We design the Self-Refined Online Foreground-Background classifier (\emph{SROFB}) to perform pixel-wise binary classification for segmentation. We train the proposed \emph{SROFB} not only with the pixels in the support images, but also with those of the query image, which bridges the semantic gap between the support and the query and facilitates the adaptation of our model to the query image.
\item We conduct extensive experiments to evaluate our method both quantitatively and qualitatively on PASCAL-5$^i$ and COCO-20$^i$ datasets, which validate the effectiveness of two novel designs described above. Furthermore, our method compares favorably with other state-of-the-art methods across all comparisons in different experimental settings.
\end{itemize}


%% file: Related_work.tex
\subsection{Semantic Segmentation}
Semantic segmentation is a fundamental task in computer vision, which aims to classify each pixel in an image. Recently its performance has been greatly improved by the end-to-end full convolutional network (FCN)~\cite{long2015fully}. The subsequent methods improve on it by proposing various modules. During the encoder stage, the resolution of the feature map keeps decreasing. To map the low resolution feature to the resolution of original imgae for pixel-wise classification, many methods~\cite{hariharan2011semantic, noh2015learning, kendall2015bayesian, yuan2020object,ronneberger2015u} propose encoder-decoder architecture.
The encoder captures higher semantic information, and the decoder recovers the spatial information.
To solve the conflict in the dense task, i.e., maintaining the feature resolution and increasing the receptive field, Chen et al.~\cite{chen2014semantic, chen2017rethinking, chen2018encoder,chen2017deeplab} propose dilated convolution, which makes it possible to maintain a higher resolution while increasing size of the receptive field during encoder stage. In our method, we also adapt the dilated convolution to maintain the resolution of feature. Beyond that, there are many modules are proposed to improve the performance of segmentation, such as deformable convolution~\cite{dai2017deformable}, contextual aggregation modules~\cite{fu2019dual, fu2019adaptive, zhao2018psanet, yuan2020object}, and multi-scale for images and features~\cite{lin2017feature, zhao2017pyramid, ghiasi2016laplacian, he2019dynamic, ding2018context, he2019adaptive}. However, the problem with these methods is that they need to rely on a large amount of labeled data, while our method mainly addresses semantic segmentation in few-shot scenarios.

\subsection{Few-shot Learning}
Few-shot learning aims to learn new concepts using few labeled data. Due to the fact that it uses only a small amount of labeled data, it has received a lot of attention in recent years. Currently there are three mainstream methods. The first method is transfer learning~\cite{Dhillon2020A, li2019large, Qi2018LowShotLW, chen2021meta, pei2022few, wu2022multi}, which consists of two main stages. The first stage uses the base categories to learn a feature encoder, and the second stage uses few data to fine-tune the classifier head. The second method is optimization-based approach~\cite{finn2017model, rusu2018meta, lee2018gradient, bertinetto2018meta}, which uses base categories to learn a good feature space as initialization. And it is able to quickly converge the model using only a small amount of data. The third is metric-based approach~\cite{vinyals2016matching, snell2017prototypical, allen2019infinite, yan2019dual,guo2022learning, cao2022learning, das2019two}, which extracts support prototypes and query prototype separately by using a siamese network and classifies the query image according to the distance between the prototype of the query and the supports. Metric-based approach is also widely used in few-shot segmentation task. In our approach, metric learning and discriminative learning are jointly used to obtain an encoder with better generalization. The metric learning is used to learn the relationship between the support images and the query image. And the discriminative learning is used for activating the discriminability of novel classes.

\subsection{Few-shot Segmentation}
Few-shot segmentation is designed to achieve segmentation of novel categories quickly using only few labeled data. Since few-shot segmentation is a task to classify dense pixels, it is more challenging than few-shot classification. Previous methods use siamese networks to solve few-shot segmentation~\cite{siam2019amp, tian2020differentiable, tian2020prior, wang2019panet, liu2020part, shaban2017one, zhang2019canet, zhang2020sg, li2020fss,liu2020crnet,fan2022self,liu2021harmonic,boudiaf2021few}. OSLSM~\cite{shaban2017one} first applies few-shot learning to the semantic segmentation task, which uses a conditional branch to generate a series of parameters from the support images as classifier. And the classifier is used to achieve the segmentation of the query image. The subsequent methods are mainly divided into non-parametric prototype-matching paradigm and parametric relation-decoding paradigm.

Inspired by ProtoNet~\cite{snell2017prototypical} in few-shot classification, many methods adapt non-parametric prototype-matching paradigm to solve few-shot segmentation task. The method of PANet~\cite{wang2019panet} first gets the foreground and background prototypes from support images, and segments the query image via computing the cosine distance between query feature at each spatial location and the prototypes obtained from the support. In contrast to the PANet~\cite{wang2019panet}, PPNet~\cite{liu2020part} and PMMs~\cite{yang2020prototype} decompose the holistic prototype representation into a set of part-aware prototypes, to capture diverse and fine-grained object features of support. It is worth noting that Self-Support Network~\cite{fan2022self} also proposes to leverage the query to improve the segmentation performance. Nevertheless, it differs from our work greatly in that it follows the prototype-matching paradigm and aims to use query to improve the quality of the prototype and thereby improves matching accuracy. By contrast, our work learns an online-classifier for segmentation, which performs self-refining using the query image.

Motivated by RelationNet~\cite{sung2018learning} in few-shot classification, many methods~\cite{zhang2019canet, tian2020prior, liu2020crnet, li2020fss} use parametric network to learn similarity relationship between support images and query image instead of using fixed measures, such as cosine similarity. CANet~\cite{zhang2019canet} extracts the holistic foreground prototype from the support image, which is concatenated with the features of query image at each spatial location. And the features are input to the dense comparison module to obtain the segmentation result of query image. Subsequently PFENet~\cite{tian2020prior} proposes feature enrichment module, which performs feature comparison at multi-scales in the feature space to adequately guide the segmentation of query image. 

Following works also propose many modules to improve performance of segmentation, including, memory-bank based~\cite{wu2021learning, xie2021few}, graph neural networks~\cite{zhang2019pyramid, wang2020few, xie2021scale} and learning classifier approaches~\cite{lu2021simpler}. However, the existing methods treat the novel classes as background during the training on the base classes, which suppresses the feature learning of novel classes, and fail to consider the underlying semantic gap between the support and query image. In this work, we aim to circumvent such limitation by activating the discriminability of novel classes in both the feature encoding stage and the prediction stage for segmentation.

%% file: Method.tex
\begin{figure*}[!t]
\centering
\includegraphics[width=1.0\linewidth]{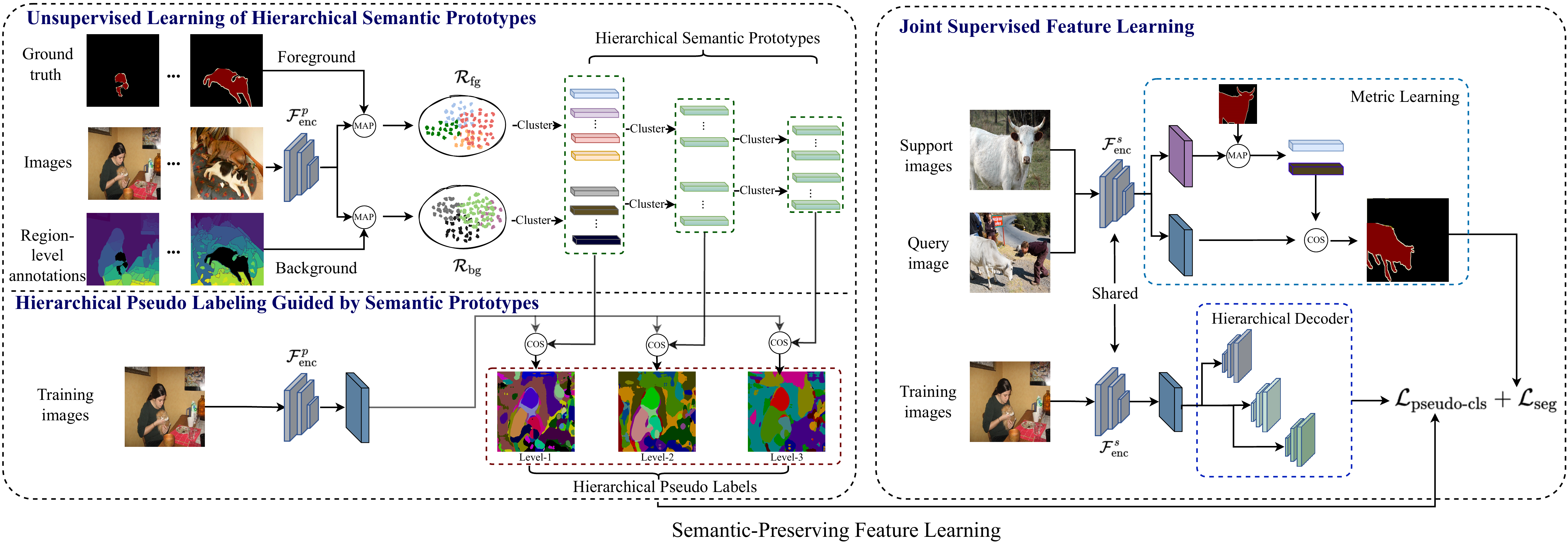}
\caption{Framework of the proposed Semantic-Preserving Feature Learning (\emph{SPFL}). It consists of three successive steps. 1) It first extracts hierarchical semantic prototypes from the entire training data by performing unsupervised region segmentation and clustering. 2) Then \emph{SPFL} leverages the obtained prototypes to conduct hierarchical pseudo labeling in pixel level for each training image; 3) Finally, \emph{SPFL} performs supervised learning of pixel-level classification using the pseudo labels to learn an effective feature encoder which can preserve the explored latent semantics.}
\label{fig:framwork}
\end{figure*}

Typical methods for few-shot segmentation treat the novel classes as background during the training on base classes, which suppresses the feature learning for objects of novel classes and impedes the segmentation of novel classes. Thus, a crucial challenge of few-shot segmentation is how to preserve the semantics of latent novel classes when performing feature learning on base classes, thereby learning discriminative features for objects of novel classes. Another critical challenge of few-shot segmentation stems from the semantic gap between the support and the query due to the scarcity of provided support samples. To deal with these two challenges, our proposed method aims to activate the discriminability of novel classes in both the feature encoding stage and the prediction stage for segmentation to achieve precise segmentation of novel classes. First, we propose the Semantic-Preserving Feature Learning (\emph{SPFL}) module to exploit and retain latent semantics contained in the whole input image, especially those in the background that may belong to novel classes, during the feature encoding for base classes. Second, we design a Self-Refined Online Foreground-Background (\emph{SROFB}) classifier for segmentation of novel classes, which is able to refine itself using the (high-confidence pixels of) query image to facilitate its adaptation to the query image. 

In this section we will first present the overall framework of the proposed method. Then we will elaborate
on the \emph{SPFL} module and \emph{SROFB} classifier, respectively.

\subsection{Overview}
The goal of few-shot segmentation is to learn effective segmentation models on a set of base classes denoted as $\mathcal{C}_\text{base}$ with sufficient samples, which can be generalized to novel classes $\mathcal{C}_\text{novel}$ with only a few labeled samples available (termed as support images). Note that there is no overlap between $\mathcal{C}_\text{base}$ and $\mathcal{C}_\text{novel}$. 
Most of existing methods focus on segmenting the target objects of base classes to learn the feature encoder during the training on base classes, whereas all other semantics are considered as background. Consequently, it is challenging for the feature encoder of such methods to learn discriminative features for novel classes, which adversely affect the segmentation of novel classes. To address this potential limitation, our proposed method is designed to activate the discriminability of novel classes in both the feature encoding stage and the prediction stage for novel-class segmentation. 

\smallskip\noindent\textbf{\emph{SPFL} for feature encoding.} We propose the Semantic-Preserving Feature Learning (\emph{SPFL}) module to learn a feature encoder which should be effective for all latent semantics including those of novel classes. To this end, the proposed \emph{SPFL} module first extracts a set of potential semantic prototypes from the entire training data in an unsupervised manner, each prototype basically representing a semantic element such as (a sub-class or super-class of) a object category or a type of background scene. Then \emph{SPFL} performs pixel-level pseudo labeling for each image in the training set based on the similarity to the obtained semantic prototypes. Finally, we leverage the obtained pseudo labels to learn the feature encoder by supervised learning of classification in pixel level. Such supervised learning of \emph{SPFL} module is performed jointly with the learning of segmentation on base classes, thus the learned feature encoder is able to preserve the discriminability for pseudo-labeled semantics including those of potential novel classes. Besides, to exploit the multi-scale latent semantics, our \emph{SPFL} produces hierarchical semantic prototypes and then obtains multi-granularity pseudo labels which allows for hierarchical supervised learning.  

\smallskip\noindent\textbf{\emph{SROFB} classifier for segmentation.} In the prediction stage for novel-class segmentation, we design the Self-Refined Online Foreground-Background (\emph{SROFB}) classifier to perform pixel-wise binary classification on the query image. To bridge the semantic gap between a query and its associated support images, we not only train the proposed \emph{SROFB} classifier using the labeled pixel data in the support images, but also refine it using (the high-confidence pixels of) the query image. To be specific, we perform a rough segmentation following the prototype-matching paradigm. Then we select both the foreground and background pixels with high confidence (matching score) in the query images as the positive and negative training samples to refine \emph{SROFB} classifier, respectively. As a result, the proposed \emph{SROFB} classifier is able to adapt to the query image effectively for segmentation and achieves more precise segmentation.

\subsection{Semantic-Preserving Feature Learning (SPFL)}
The proposed \emph{SPFL} module consists of three successive steps, as illustrated in Figure~\ref{fig:framwork}. It first extracts hierarchical semantic prototypes from the entire training data in an unsupervised way to explore multi-scale latent semantics. Then the obtained semantic prototypes are used to conduct pixel-level pseudo labeling for each image in the training set by assigning each pixel to the semantically nearest prototype. Finally, \emph{SPFL} performs supervised learning of pixel-wise classification using the pseudo labels to learn an effective feature encoder which can preserve the explored latent semantics. 

\subsubsection{Unsupervised Learning of Hierarchical Semantic Prototypes}
In the problem setting of few-shot segmentation, only the samples of base classes and the associated mask annotations of foreground objects in base classes are available during the training phase. To explore other latent semantics in the background area, especially those sharing little semantic similarities with base classes, we perform semantic region segmentation for the background area of each image in the training set in an unsupervised way. Then we cluster these segmented regions to produce the representative semantic prototypes by calculating the clustering centers.  

\smallskip\noindent\textbf{Unsupervised region segmentation.} We adopt MCG~\cite{arbelaez2014multiscale}, which is a prominent non-parametric super-pixel segmentation method, to perform unsupervised region segmentation considering its superior performance. MCG is able to detect contours in an image efficiently based on appearance information like brightness and texture. We employ MCG to obtain a fine-grained region segmentation for the background area of each image in the training set by tuning a proper segmentation threshold.

As shown in Figure~\ref{fig:framwork}, we employ a pre-trained feature encoder $\mathcal{F}_\text{enc}^p$ to learn features for an input image and and then crop the feature for each segmented region in the image according to the segmented mask. The feature encoder $\mathcal{F}_\text{enc}^p$ is built as a ResNet-50 or ResNet-101~\cite{he2016deep} network which is pre-trained on ImageNet~\cite{deng2009imagenet} and fine-tuned on target datasets following the prototype-matching paradigm. Then we extract the vectorial representation for each segmented region by Mask Average Pooling (MAP)~\cite{siam2019amp}. For instance, the vectorial representation $\mathbf{p}_{i}^{j}$ for the $j$-th region in the $i$-th image is calculated by:
\begin{equation} 
    \mathbf{p}_{i}^{j}=\frac{\sum_{u, v} \mathbf{F}_{i}^{u, v} \mathbbm{1}\left[M_{i}^{r(u, v)}=j\right]}{\sum_{u, v} \mathbbm{1}\left[M_{i}^{r(u, v)}=j\right]},
\label{eqn:map}
\end{equation}
where $\mathbf{F}_{i}^{u, v}$ denotes the learned features by the feature encoder $\mathcal{F}_\text{enc}^p$ for the pixel located at $(u,v)$ in the $i$-th image and $M_{i}^{r(u, v)}=j$ indicates the pixels of the $j$-th region in the $i$-th image. Consequently, we can collect a corpus of segmented regions from the background area of all training images, which we denote as $\mathcal{R}_\text{bg}$. Meanwhile, we can easily achieve a corpus of foreground objects from the training set, denoted as $\mathcal{R}_\text{fg}$, based on the provided mask annotations of objects.

\smallskip\noindent\textbf{Hierarchical clustering for producing prototypes.} We perform clustering on the corpus of segmented background regions $\mathcal{R}_\text{bg}$ as well as the corpus of foreground objects $\mathcal{R}_\text{fg}$, and derive a set of latent semantic prototypes by calculating the vectorial representations of the cluster centers. Specifically, we apply K-means algorithm to cluster $\mathcal{R}_\text{fg}$ and $\mathcal{R}_\text{bg}$ respectively:
\begin{equation}
\begin{split}
    &\mathcal{P}_\text{fg} = \text{K-means}(\mathcal{R}_\text{fg}; K_\text{fg}),\\
    &\mathcal{P}_\text{bg} = \text{K-means}(\mathcal{R}_\text{bg}; K_\text{bg}),
        \end{split}
\label{eqn:cluster}
\end{equation}
where $\mathcal{P}_\text{fg}$ and $\mathcal{P}_\text{bg}$ are the sets of derived foreground and background prototypes, which contains $K_\text{fg}$ and $K_\text{bg}$ cluster centers, respectively.

To exploit multi-scale latent semantics, we perform hierarchical clustering to obtain hierarchical semantic prototypes. To be specific, we first cluster the foreground and background regions to derive  fine-grained prototype sets, as shown in Equation~\ref{eqn:cluster}. Then we perform iterative clustering on the prototype sets obtained in the previous clustering step with a smaller number of clusters ($K_\text{fg}$ and $K_\text{bg}$) to derive a coarser prototype set, in which each prototype corresponds to a superclass of a cluster of finer-grained prototypes in the previous step. As a result, we are able to derive coarse-to-fine hierarchical semantic prototypes (containing 3 levels of prototype sets in our implementation) for the foreground and background areas, respectively.

\subsubsection{Hierarchical Pseudo Labeling guided by Semantic Prototypes}
We leverage the obtained hierarchical semantic prototypes to perform pixel-wise pseudo labeling. 
Specifically, each semantic prototype can be considered as a class center. We measure the Cosine similarity between a pixel to each prototype in the feature space and assign it to the prototype that is semantically closest to the pixel. Note that the foreground and background pixels are labeled using the foreground prototypes $\mathcal{P}_\text{fg}$ and the background prototypes $\mathcal{P}_\text{bg}$, respectively. 

Since different level of prototypes corresponds to different scale of latent semantics, we perform pseudo labeling using each level of prototypes separately and derive multi-granularity pseudo labels for each pixel, as shown in Figure~\ref{fig:framwork}. Formally, a background pixel located at $(u,v)$ in the $i$-th image is labeled in the $l$-th level of granularity by:
\begin{equation}
    \mathbf{y}^{u,v}_{i, l} = \underset{k}{\arg \max } \cos(\mathbf{F}^{u,v}_i, \mathcal{P}_{\text{bg}, l}^k),
\end{equation}
where $\mathcal{P}_{\text{bg}, l}^k$ is the $k$-th prototype in the $l$-th level of prototype sets.

\subsubsection{Joint Supervised Feature Learning}
The obtained pixel-level pseudo labels are further used for supervised learning of feature encoder for segmentation $\mathcal{F}_\text{enc}^s$ by pixel-wise classification, which is a pretext task designed to enable $\mathcal{F}_\text{enc}^s$ to preserve the latent semantics characterized by the semantic prototypes during the training phase for base classes. Note that the feature encoder for segmentation is learned independently from the pre-trained feature encoder $\mathcal{F}_\text{enc}^p$. We apply the cross-entropy loss ($\mathcal{L}_\text{CE}$) to perform supervised learning for each pixel in all training images,  with each level of hierarchical pseudo labels used individually. Formally, the supervision using the $l$-th level of pseudo labels is performed by:
\begin{equation}
    \mathcal{L}_{\text{pseudo-cls}, l} = \sum_{i=1}^{N}\sum_{u=1,v=1}^{H,W} \mathcal{L}_\text{CE}(\mathbf{y}^{u,v}_{i,l}, \mathcal{F}_\text{dec}^l(\mathbf{F}_i^{u,v})),
\end{equation}
where $\mathbf{F}_i^{u,v}$ is the extracted features from the feature encoder $\mathcal{F}_\text{enc}^s$ for the pixel at $(u, v)$ in the $i$-th image, which is further fed into a decoder $\mathcal{F}_\text{dec}^l$ for feature decoding and pixel-wise prediction. It consists of two $3\times 3$ convolutional layers and one $1\times 1$ convolutional layer plus a bilinear upsampling layer projecting the decoded feature maps to the same size as the input image. Note that we learn independent feature decoders for different levels of pseudo labels, which have the same model structure but independent parameters. $\mathbf{y}^{u,v}_{i,l}$ is the associated pseudo label in the $l$-th level of granularity. $N, (H, W)$ are the amount of training images and the image size (height and width), respectively. The overall supervision using different levels of pseudo labels is performed as the weighted sum of the losses for each level:
\begin{equation}
    \mathcal{L}_{\text{pseudo-cls}} = \sum_{l=1}^L \gamma_l\mathcal{L}_{\text{pseudo-cls}, l},
    \label{eqn:hierarchy}
\end{equation}
where $\gamma_l$ is a hyper-parameter denoting the weight for the supervision loss of the $l$-th level of pseudo labels.

Apart from the supervision using the pseudo labels for learning the feature encoder $\mathcal{F}_\text{enc}^s$, we also perform support-query metric learning for segmentation of base classes with the provided groundtruth masks, following the typical prototype-matching paradigm~\cite{siam2019adaptive,wang2019panet,liu2020part,yang2021mining}. To be specific, we classify each pixel of the query image as foreground or background according to the Cosine similarities in the encoded feature space between it and the foreground and background prototypes from the support images. Thus, the segmentation can be formulated as a binary classification task supervised by a binary cross-entropy loss $\mathcal{L}_\text{BCE}$: 
\begin{equation} \label{eq4}
    \mathcal{L}_\text{seg}= \sum_{i=1}^{N}\sum_{u=1,v=1}^{H,W} \mathcal{L}_\text{BCE}(\mathbf{z}^{u,v}_{i}, \mathbf{s}^{u,v}_{i}),
\end{equation}
where $\mathbf{z}^{u,v}_{i}$ is the binary groundtruth label for the pixel at $(u,v)$ in the $i$-th image. $\mathbf{s}^{u,v}_{i}$ is the paired Cosine similarities between the pixel and the foreground prototype $\mathbf{p}_\text{fg}$ and background prototype $\mathbf{p}_\text{bg}$: 
\begin{equation}
    \mathbf{s}^{u,v}_{i} = [\cos(\mathbf{F}_i^{u,v}, \mathbf{p}_\text{bg}); \cos(\mathbf{F}_i^{u,v}, \mathbf{p}_\text{fg})].
    \label{eqn:sim}
\end{equation}
Herein, the foreground prototype $\mathbf{p}_\text{fg}$ and background prototype $\mathbf{p}_\text{bg}$ are calculated by Mask Average Pooling (MAP) (shown in Equation~\ref{eqn:map}) over the foreground and background of the supported images according to the groundtruth masks respectively, which are different from the semantic prototypes obtained by clustering.


We learn the feature encoder for segmentation $\mathcal{F}_\text{enc}^s$ in such a joint supervised learning manner that the trained feature encoder is able to not only learn a good similarity measure between query images and their associated support images, but also exploit and preserve the latent semantics, especially those contained in the background area. As a result, our model can learn discriminative features for novel classes during feature encoding, which is crucial for the segmentation of novel classes.

\begin{figure*}[!t]
\centering
\includegraphics[width=1.0\linewidth]{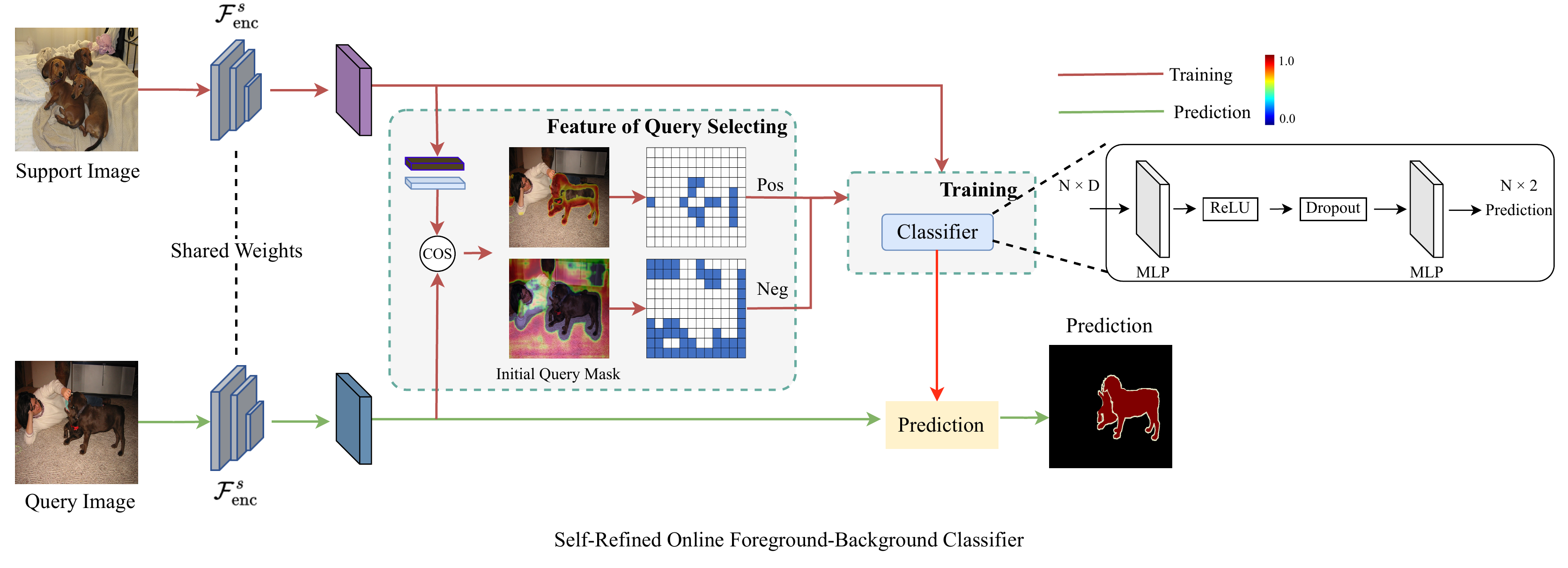}
\caption{Illustration of the proposed Self-Refined Online Foreground-Background classifier (\emph{SROFB}). It is not only trained with the pixels of support images, but also refined using the high-confidence pixels of the query image. It first perform a rough segmentation on the query image following the prototype-matching paradigm between the support and the query, then we select reliable pixels with high matching score from both the foreground and background in the query as the positive and negative training samples for \emph{SROFB}, respectively. As a result, \emph{SROFB} is able to adapt to query image smoothly and bridge the semantic gap between the support and the query.}
\label{fig:SROFB}
\end{figure*}
\subsection{Self-Refined Online Foreground-Background Classifier (SROFB)}
Typical methods based on prototype-matching paradigm segment novel classes by measuring the semantic similarities between each pixel of the query image and both the foreground and background prototypes derived from the support images, performing in the learned feature encoding space on base classes. A potential downside of such methods is that the underlying semantic gap between the query and the associated support images, typically arising in few-shot learning scenario, limits the matching accuracy and the segmentation performance. To circumvent this challenge, we learn an Online Foreground-Background classifier (\emph{SROFB}), illustrated in Figure~\ref{fig:SROFB}, which performs pixel-wise binary classification for segmentation. We not only train \emph{SROFB} classifier using the (pixels of) support images, but also refine it with the query image to adapt the classifier to the query image seamlessly and thus bridge the query-support semantic gap.

We can readily obtain the positive and negative pixels for training the \emph{SROFB} classifier from the support images based on the groundtruth mask annotations. For the query image whose mask is not available, we first perform a rough segmentation on the query image following the prototype-matching paradigm between support images and the query. Then we select the reliable pixels with high matching score from both the foreground and background in the query as the positive and negative training samples for \emph{SROFB} respectively, as illustrated in Figure~\ref{fig:SROFB}. Specifically, the matching score $\mathbf{c}_\text{q}^{u,v}$ of the pixel at $(u, v)$ in a query image denoted as $q$ is calculated by applying Softmax function to the paired similarities $\mathbf{s}_\text{q}^{u,v}$ between the pixel and the foreground prototype and the background prototype:
\begin{equation}
    \mathbf{c}_\text{q}^{u,v} = \text{Softmax}(\mathbf{s}_\text{q}^{u,v}),
\end{equation}
where $\mathbf{s}_\text{q}^{u,v}$ is calculated by Equation~\ref{eqn:sim}. The resulting matching score $\mathbf{c}_\text{q}^{u,v} \in \mathrm{R}^2$ comprises the foreground matching score and the background matching score. We preset two thresholds $\tau_\text{fg}$ and $\tau_\text{bg}$ for the foreground and background matching scores respectively, to select the positive and negative samples for training \emph{SROFB}. Consequently, the selected training pixels from the query and the support pixels compose the training set for learning the \emph{SROFB} classifier.

As shown in Figure~\ref{fig:SROFB}, we design the \emph{SROFB} classifier as a lightweight neural network, which consists of two Multi-perception layers with a ReLU layer and a Dropout layer in between for nonlinear transformation. Thus, \emph{SROFB} can be trained quite efficiently in an online fashion with the binary cross-entropy loss. Compared to other paradigms for segmentation such as the prototype-matching paradigm~\cite{siam2019adaptive,wang2019panet, yang2021mining,liu2020part, lu2021simpler} or the parametric relation-decoding paradigm~\cite{zhang2019canet, yang2020prototype, liu2020crnet, tian2020prior, li2020fss}, another prominent advantage of the proposed \emph{SROFB} classifier is that it has better scalability with the increase of support images. It can achieve more performance gain than other segmentation paradigms when given more annotated supported images, which is experimentally validated in Section~\ref{sec:ablation}.

%% file: Experiment.tex
\begin{table*}[!t]
\renewcommand{\arraystretch}{1.2}
    \caption{Quantitative comparison results on PASCAL-5$^i$ dataset. $\dagger$ indicates the reproduced results using the released code while other results are reported in other papers.}
    \label{table_voc}
    \centering
\resizebox{\linewidth}{!}{
\begin{tabular}{l|c|ccccc|ccccc|c}
\toprule
\multirow{2}{*}{Method} & \multirow{2}{*}{Backbone} & \multicolumn{5}{c|}{1-shot} & \multicolumn{5}{c|}{5-shot} & \multirow{2}{*}{Params} \\
 & & fold0 & fold1 & fold2 & fold3 & \textbf{Mean} &  fold0  &  fold1  & fold2   &  fold3  &  \textbf{Mean} &                   \\ \midrule
PANet~\cite{wang2019panet} (ICCV’19)& \multirow{9}{*}{ResNet-50}  &  44.0 &57.5 & 50.8  & 44.0  & 49.1  & 55.3  & 67.2 & 61.3 &  53.2  & 59.3 & 23.5M \\ 
CANet~\cite{zhang2019canet} (CVPR’19)  & & 52.5 &65.9 & 51.3  & 51.9  & 55.4  & 55.5  & 67.8 & 51.9 &  53.2  & 57.1 &36.4M \\ 
PPNet~\cite{liu2020part} (ECCV’20) &       & 48.6  & 60.6  & 55.7  & 46.5  & 52.8  & 58.9  & 68.3  & 66.8  & 58.0    & 63.0    & \multicolumn{1}{c}{31.5M} \\
PMMs~\cite{yang2020prototype} (ECCV’20) &       & 55.2  & 66.9  & 52.6  & 50.7  & 56.3  & 56.3  & 67.3  & 54.5  & 51.0    & 57.3  & \multicolumn{1}{c}{19.6M} \\
PFENet~\cite{tian2020prior} (TPAMI’20) &       & 61.7  & 69.5  & 55.4  & \textbf{56.3}  & 60.8  & 63.1  & 70.7  & 55.8  & 57.9  & 61.9  & \multicolumn{1}{c}{34.3M} \\
RePRI~\cite{boudiaf2021few} (CVPR’21) &       & 59.8  & 68.3    & 62.1 & 48.5  & 59.7  & 64.6  & 71.4  & 71.1  & \textbf{59.3}  & 66.6  & \multicolumn{1}{c}{46.7M} \\
CWT~\cite{lu2021simpler} (ICCV’21) &       & 56.3  & 62.0    & 59.9  & 47.2  & 56.4  & 61.3  & 68.5  & 68.5  & 56.6  & 63.7  & \multicolumn{1}{c}{48.8M}  \\
CWT$^\dagger$~\cite{lu2021simpler} (ICCV’21) &       & 55.9  & 61.4  & 58.8  & 46.8  & 55.7    & 61.8  & 66.4  & 64.6  & 53.4  & 61.5  & \multicolumn{1}{c}{48.8M}  \\
HFA~\cite{liu2021harmonic} (TIP’2021) &       & 53.0  & 69.0  & 53.5  & 51.7  & 56.8    & 55.2  & 70.0  & 56.8  & 52.6  & 58.7  & \multicolumn{1}{c}{$-$}   \\
MLC~\cite{yang2021mining} (ICCV’21) &       & 59.2  & \textbf{71.2}  & 65.6  & 52.5  & 62.1  & 63.5  & 71.6  & 71.2  & 58.1  & 66.1  & \multicolumn{1}{c}{8.7M} \\
Ours  &       & \textbf{61.8}  & 69.3  & \textbf{71.0}  & 52.7    & \textbf{63.7}  & \textbf{64.6}  & \textbf{72.3}  & \textbf{74.0}  & 58.8  & \textbf{67.4}  & 9.0M \\ \midrule \midrule
FWB~\cite{nguyen2019feature} (ICCV’19) &  \multirow{8}{*}{ResNet-101}      & 51.3  & 64.5  & 56.7  & 52.2  & 56.2  & 54.8  & 67.4  & 62.2  & 55.3  & 59.9  & \multicolumn{1}{c}{43.0M} \\
PPNet~\cite{liu2020part} ECCV’20) &       & 52.7  & 62.8  & 57.4  & 47.7  & 55.2  & 60.3  & 70.0    & 69.4  & 60.7  & 65.1  & \multicolumn{1}{c}{50.5M} \\
DAN~\cite{wang2020few} (ECCV’20) &       & 54.7  & 68.6  & 57.8  & 51.6  & 58.2  & 57.9  & 69.0    & 60.1  & 54.9  & 60.5  &  \multicolumn{1}{c}{$-$} \\
PFENet~\cite{tian2020prior} (TPAMI’20) &       & 60.5  & 69.4  & 54.4  & 55.9  & 60.1  & 62.8  & 70.4  & 54.9  & 57.6  & 61.4  & \multicolumn{1}{c}{53.4M} \\
RePRI~\cite{boudiaf2021few} (CVPR’21) &       & 59.6  & 68.6  & 62.2  & 47.2  & 59.4  & \textbf{66.2}  & 71.4  & 67.0    & 57.7  & 65.6  & \multicolumn{1}{c}{65.7M} \\
CWT~\cite{lu2021simpler} (ICCV’21) &       & 56.9  & 65.2  & 61.2  & 48.8  & 58.0    & 62.6  & 70.2  & 68.8  & 57.2  & 64.7  & \multicolumn{1}{c}{67.8M}   \\
CWT$^\dagger$~\cite{lu2021simpler} (ICCV’21) &       & 56.4  & 63.9  & 59.8  & 47.2  &   56.8  & 63.4  & 68.8 & 67.3  & 54.6  & 63.5  & \multicolumn{1}{c}{67.8M}   \\
MLC~\cite{yang2021mining} (ICCV’21) &       & \textbf{60.8}  & \textbf{71.3}  & 61.5  & \textbf{56.9}  & 62.6  & 65.8  & 74.9  & 71.4  & \textbf{63.1}  & 68.8  & \multicolumn{1}{c}{27.7M} \\
Ours  &       & 60.5 & 70.5 & \textbf{71.0} & 54.7 & \textbf{64.2} & 64.6    & \textbf{75.3}    & \textbf{77.1} & 61.1 & \textbf{69.5} & 28.0M  \\

\bottomrule
\end{tabular}}
\end{table*}

We conduct extensive experiments to evaluate our model both quantitatively and qualitatively. Specifically, we first compare our model to other state-of-the-art methods for few-shot semantic segmentation following standard evaluation protocols,  then we conduct detailed ablation study to investigate the effectiveness of each proposed techniques.
\subsection{Dataset and Evaluation metric}
\subsubsection{Dataset}
In our experience, we evaluate our method on two datasets, including PASCAL-$5^i$ and COCO-$20^i$. The PASCAL-$5^i$~\cite{shaban2017one} is the extension of PASCAL VOC 2012~\cite{everingham2010pascal} and augmented SBD~\cite{hariharan2011semantic} which contains 20 categories. Following OSLSM~\cite{wang2019panet}, we divide the 20 categories into 4 splits $i \in \{0, 1, 2, 3\}$, and each split contains 5 categories. Following the standard evaluation protocol for few-shot segmentation~\cite{shaban2017one, wang2019panet}, when conducting each experiment, cross-validation is performed. Each time three out of four splits are viewed as the base classes for training, while the remaining split is used as the novel categories for test. Therefore, we can construct four experiments, denoted as \textit{fold0, fold1, fold2} and  \textit{fold3}. COCO-$20^i$~\cite{nguyen2019feature} is a challenging dataset modified from COCO~\cite{lin2014microsoft}. It contains 80 categories and quite complex scenes. Following FWB~\cite{nguyen2019feature}, the 80 categories are divided into 4 splits, each with 20 categories. Cross validation is conducted, which is similar to the experimental setting of PASCAL-$5^i$.

\begin{table*}[!h]
\renewcommand{\arraystretch}{1.2}
    \caption{Quantitative comparison results on COCO-5$^i$ dataset. $\dagger$ indicates the reproduced results using the released code while other results are reported in other papers.}
    \label{table_coco}
    \centering
\resizebox{\linewidth}{!}{
\begin{tabular}{l|c|ccccc|ccccc|c}
\toprule
\multirow{2}{*}{Method} & \multirow{2}{*}{Backbone} & \multicolumn{5}{c|}{1-shot} & \multicolumn{5}{c|}{5-shot} & \multirow{2}{*}{Params} \\
 & & fold0 & fold1 & fold2 & fold3 & \textbf{Mean} &  fold0  &  fold1  & fold2   &  fold3  &  \textbf{Mean} &                   \\ \midrule
PANet~\cite{wang2019panet} (ICCV’19) &   \multirow{5}{*}{ResNet-50}     & 31.5  & 22.6  & 21.5  & 16.2  & 23.0  & 45.9    & 29.2  & 30.6  & 29.6  & 33.8  & \multicolumn{1}{c}{23.5M} \\
PPNet~\cite{liu2020part} (ECCV'20) &       & 36.5  & 26.5  & 26.0    & 19.7  & 27.2  & 48.9  & 31.4  & 36.0    & 30.6  & 36.7  & \multicolumn{1}{c}{31.5M} \\
CWT~\cite{lu2021simpler} (ICCV'21)   &       & 32.2  & \textbf{36.0}    & \textbf{31.6}  & \textbf{31.6}  & 32.9  & 40.1  & \textbf{43.8}  & \textbf{39.0}    & \textbf{42.4}  & \textbf{41.3}  & \multicolumn{1}{c}{48.8M} \\
CWT$^\dagger$~\cite{lu2021simpler} (ICCV'21)   &       & 30.2  & 32.1    & 27.0  & 28.2  & 29.4  & 34.7  & 38.8  & 31.4    & 32.2  & 34.3  &\multicolumn{1}{c}{48.8M}   \\
HFA~\cite{liu2021harmonic} (TIP'21)   &       & 27.5  & 35.0  & 29.2  & 32.2  & 31.0  & 31.5  & 41.0  & 28.5  & 34.9  & 34.0  & \multicolumn{1}{c}{$-$} \\
MLC~\cite{yang2021mining} (ICCV'21)   &       & 46.8  & 35.3  & 26.2  & 27.1  & 33.9  & 54.1  & 41.2  & 34.1  & 33.1  & 40.6  & \multicolumn{1}{c}{8.7M} \\
Ours  &       & \textbf{47.1}  & 35.2  & 26.3  & 27.3  & \textbf{34.0}  & \textbf{54.8}  & 40.9  & 34.3  & 33.0    & 40.8 &9.0M \\   \midrule \midrule
PMMs~\cite{yang2020prototype} (ECCV'20)  &  \multirow{4}{*}{ResNet-101}     & 29.5  & 36.8  & 28.9  & 27.0    & 30.6  & 33.8  & 42.0    & 33.0    & 33.3  & 35.5  & \multicolumn{1}{c}{38.6M} \\
CWT~\cite{lu2021simpler} (ICCV'21)  &       & 30.3  & 36.6  & \textbf{30.5}  & \textbf{32.2}  & 32.4  & 38.5  & \textbf{46.7}  & \textbf{39.4}  & \textbf{43.2}  & 42.0    & \multicolumn{1}{c}{67.8M}   \\
MLC~\cite{yang2021mining} (ICCV'21)   &       & 50.2  & 37.8  & 27.1  & 30.4  & 36.4  & 57.0    & 46.2  & 37.3  & 37.2  & 44.4  & \multicolumn{1}{c}{27.7M} \\
Ours  &  & \textbf{50.3}  & \textbf{38.8}  & 28.6  & 30.6  & \textbf{37.1}      &  \textbf{57.3}  & 46.5  & 37.8  & 38.0    & \textbf{44.9} & 28.0M \\
\bottomrule
\end{tabular}}
\end{table*}

\subsubsection{Evaluation metric}
Following \cite{shaban2017one,wang2019panet, liu2020part}, we adopt popular mean Intersection-over-Union (mIoU) for performance evaluation. In the inference stage, we randomly sample 1000/4000 support-query pairs to conduct evaluation for PASCAL-5$^i$/COCO-20$^i$, respectively. By default, all ablation studies are conducted on PASCAL-$5^i$ with ResNet-50 backbone in the 1-shot setting.

\subsection{Implementation Details}

\subsubsection{Semantic-Preserving Feature Learning (SPFL)} 
To get the fine-grained region segmentation for the background area of each image in the training set, we adop MCG\cite{arbelaez2014multiscale} method and set the contour threshold $\tau=0.45$ in all experiments. Given the pre-trained backbone, which is fine-tuned on the base classes from training set using prototype-matching paradigm, called Baseline model, we extract the feature representation of the foreground object, denoted as $\mathcal{R}_\text{fg}$ and the representation of segmented regions from the background area, denoted as $\mathcal{R}_\text{bg}$, respectively. And we use the K-means algorithm on the corpus of segmented background regions $\mathcal{R}_\text{bg}$ as well as the corpus of foreground objects $\mathcal{R}_\text{fg}$ to obtain the latent semantic prototypes. To get the hierarchical prototypes, we use three levels of clustering. The number of cluster center is $\{50, 25, 15\}$ on PASCAL-$5^i$ and the number of cluster center is $\{75, 50, 25\}$ on COCO-$20^i$ according to the statistics of categories on these datasets. 

We adopt the ResNet-50/101~\cite{he2016deep} pretrained on ImageNet~\cite{deng2009imagenet} as the backbone network, respectively. The backbone we use has the same structure as the one used by MLC~\cite{yang2021mining} for better generalization, in which the last stage and the last ReLU of penultimate stage are discarded. The three decoders have the same structure, and each decoder consists of two $3\times 3$ convolutional layers and one $1\times 1$ convolutional layer, 
where each convolutional layer is followed by BN(batch normalization) and ReLU except the last one.

Given the groundtruth masks and pseudo labels, we train our model according to the following settings. On the PASCAL-$5^i$ and COCO-$20^i$, we use 4 support-query pairs and 16 extra training images as mini-batch. We use 8 extra training images in 5-shot on ResNet-101. We use SGD optimizer for training. The learning rate is initialized by 1e-3 and decays by 10 times every 2000 iterations. The weight decay is 1e-4, and the momentum is 0.9. The model is trained for 6,000 iterations. Images and masks are resized into (473, 473) for training. The images for metric learning are only augmented with random horizontal flipping, while the images with pseudo labels use a strong data augmentation following the practice in~\cite{chen2020simple}. The weights $\gamma_1$, $\gamma_2$ and $\gamma_3$ of loss $\mathcal{L}_{\text{pseudo-cls}}$ are set to 0.5, 1.0 and 1.0 for preserving the latent semantics characterized with the pseudo labels. 

\subsubsection{Self-Refined Online Foreground-Background Classifier (SROFB)}
We use the backbone parameters after training by the \emph{SPFL} as initialization and freeze it when getting the foreground-background classifier. We adapt an online manner to adjust the classifier to each testing task. First, we can get a rough segmentation on the query image following the prototype-matching paradigm between support images and the query. Then the positive and negative samples are obtained from the query with the foreground threshold $\tau_{fg}=0.7$ and the background threshold $\tau_{bg}=0.6$. The architecture of classifier is two MLP layers with ReLU and dropout in between. we use SGD as the optimizer. The learning rate is set to 0.1, and 10 iterations in 1-shot setting and 100 iterations in 5-shot setting. 

\subsection{Comparison with State-of-the-Arts}
We compare our method to state-of-the-art methods for few-shot segmentation on two benchmark datasets~\cite{lin2014microsoft,everingham2010pascal,hariharan2011semantic} using different backbones including ResNet-50 and ResNet-101, in two different few shot settings including 1-shot and 5-shot settings. 

\subsubsection{PASCAL-$5^i$}
As shown in Table~\ref{table_voc}, our method outperforms the previous methods in terms of `Mean' performance in both 1-shot and 5-shot settings using either ResNet-50 or ResNet-101 as the backbone. Specifically, our method achieves $1.6\%$ and $1.6\%$ performance gain over the state-of-the-art performance (obtained by MLC~\cite{yang2021mining}) in the 1-shot setting on Resnet-50 and Resnet-101, respectively. More importantly, in fold2 split, which contains \textit{dining table, dog, horse, motorbike} and \textit{person} for novel categories, our method outperforms other methods by at least $9.5\%$ on Resnet-101 in 1-shot setting. We observe that these categories often appear in the background of the training set (base classes), which demonstrates the advantages of our model in activating the discriminability of novel classes, especially those treated as background. In 5-shot setting, our method also performs well and compares favorably with other methods. What's more, our model, as well as MLC, have much smaller model size than other methods, which reveals another merit of our model.

\subsubsection{COCO-$20^i$}
It is a more challenging dataset than PASCAL-$5^i$ dataset due to more categories and more complex scenes. As shown in Table~\ref{table_coco}, our approach achieves best performance in 1-shot setting in terms of `Mean' metric, when using ResNet-50 as the backbone. In the 5-shot setting, our approach performs better than other methods except CWT. However, we re-evaluate CWT using the released code and the reproduced result is only $34.3\%$, which is substantially lower than the reported result ($41.3\%$). 
Using ResNet-101 as the backbone, our model achieves the best `Mean' performance in both 1-shot and 5-shot settings, which manifests the robust performance of our model on this challenging dataset with complex scenes.

\begin{figure}[!t]
\centering
\includegraphics[width=1.0\linewidth]{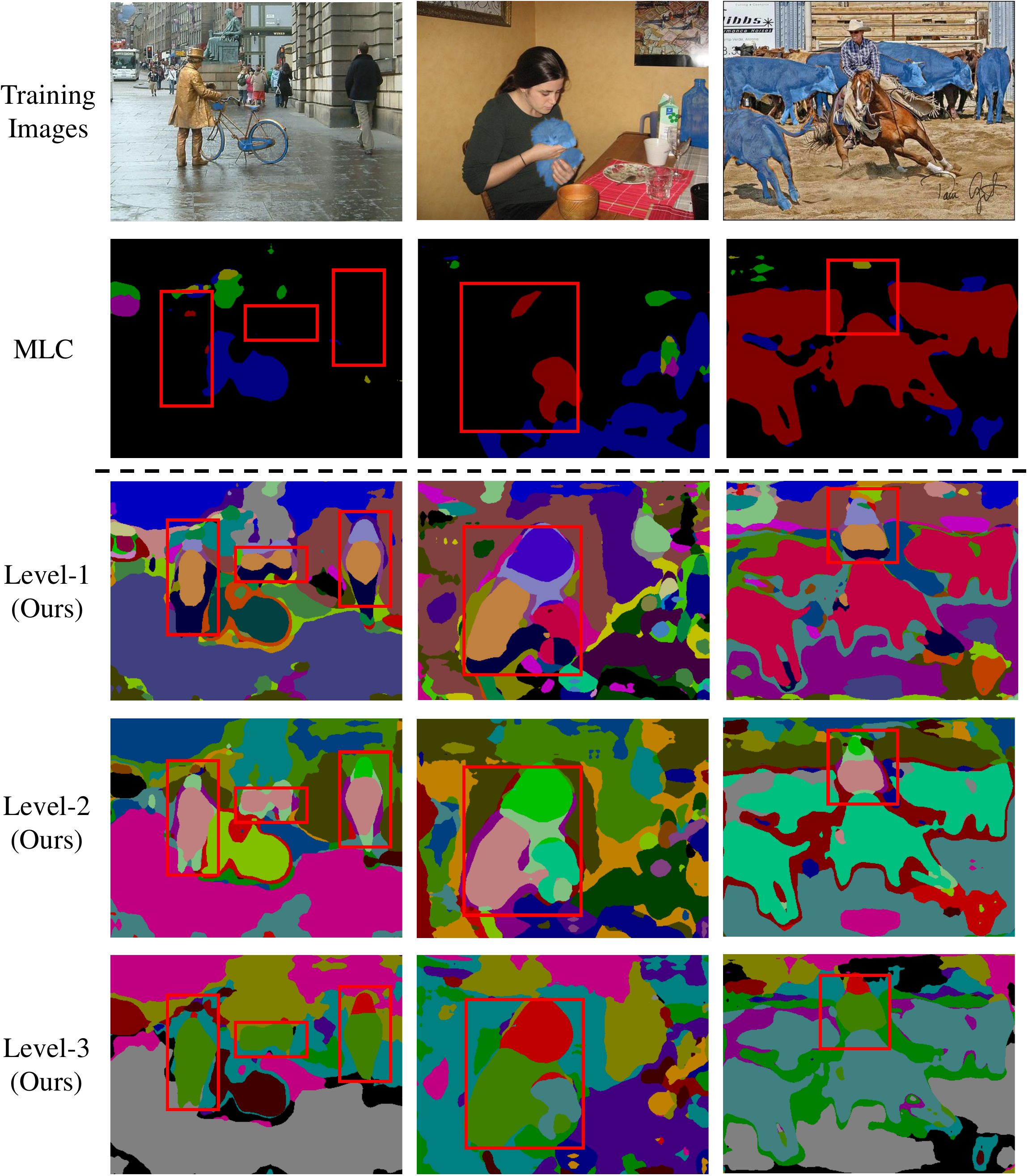}
\caption{Visualization of pseudo labeling by our method and MLC~\cite{yang2021mining}. The first row shows the initial input images with blue masks indicating the objects of base classes. The results of pseudo labeling by MLC are presented in the second row while the three levels of hierarchical pseudo labeling by our method are visualized in other rows.}
\label{fig:pseudo_vis}
\end{figure}

\subsubsection{Qualitative Evaluation}
Similar to our method, MLC~\cite{yang2021mining} also aims to explore the latent semantics in the background. Nevertheless, there are two major differences between MLC and our model: 1) MLC aims to mine the latent semantics in the background which are similar to the base classes, whilst our model is able to exploit all kinds of latent semantics in the whole image by extracting hierarchical semantic prototypes from the entire training data in an unsupervised way. 2) MLC performs segmentation following the typical prototype-matching paradigm while our model learns an online foreground-background classifier (\emph{SROFB}) for segmentation that can refine itself using the high-confidence pixels of the query image to bridge the semantic gap between the support and the query. We perform two sets of qualitative evaluation to compare our model and MLC. 

\smallskip\noindent\textbf{Comparison of pseudo labeling.} We first visualize the results of pseudo labeling by MLC and our model in Figure~\ref{fig:pseudo_vis}. It can be observed that our model is able to explore the latent novel classes in the background that are distinctly different from the base classes while MLC cannot, which validate the above analysis and demonstrates the merit of our model over MLC.

\begin{figure}[!t]
\centering
\includegraphics[width=1.0\linewidth]{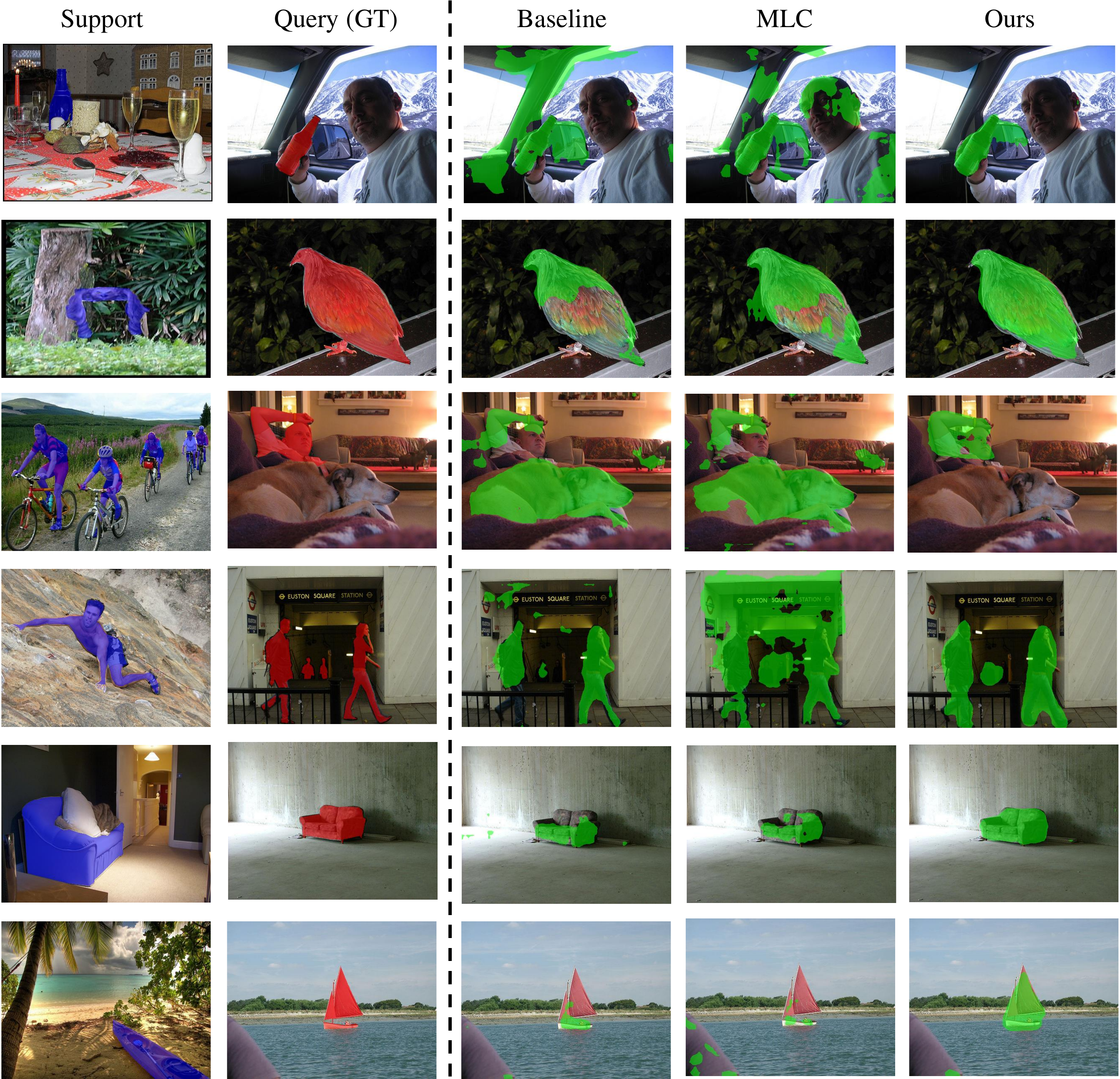}
\caption{Segmentation results by the Baseline model, MLC and our method. Herein, the baseline model is a simplified variant of MLC by removing the function of latent semantic mining.}
\label{fig:vis_segmentation}
\end{figure}

\smallskip\noindent\textbf{Comparison of segmentation.}
In the second set of qualitative evaluation, we compare the segmentation results between MLC and our method. As shown in Figure~\ref{fig:vis_segmentation}, while MLC performs slightly better than the baseline which has no capability of latent semantic mining, our model can segment the target objects much more precisely than MLC. In particular, when the object appearance of the query image differs largely from that of the support images, as shown in the `boat' example of the last row in Figure~\ref{fig:vis_segmentation}, MLC can hardy perform segmentation correctly while our method can.

\begin{table}[!t]
\centering
\renewcommand{\arraystretch}{1.2}
    \caption{Ablation study on each core functional component. The baseline model is constructed by removing \emph{SPFL} and \emph{SROFB}, which follows the standard prototype-matching paradigm for few-shot segmentation.}
    \label{table_component}
\resizebox{\linewidth}{!}{
\LARGE
\begin{tabular}{c | c c c | c c c c l}
\toprule
Baseline & \makecell[c]{Single-level \emph{SPFL}\\(w/o Hierarchy)} & \makecell[c]{\emph{SPFL}\\(w/ Hierarchy)} & \emph{SROFB} & fold0 & fold1 & fold2 & fold3 & \textbf{Mean} \\ \midrule
\checkmark & & & & 56.4  & 66.4  & 60.6  & 47.7  & 57.8 \\
\checkmark & \checkmark & & & 55.6  & 69.3  & 67.3  & 50.1  & $60.6_{\uparrow 2.8}$ \\
\checkmark& & \checkmark & &58.2  & \textbf{70.0}    & 67.1  & 50.9  & $61.6_{\uparrow 3.8}$ \\
\checkmark& &\checkmark & \checkmark &\textbf{61.8}  & 69.3  & \textbf{71.0}  & \textbf{52.7}    & $\textbf{63.7}_{\uparrow 5.9}$ \\
\bottomrule
\end{tabular}}
\end{table}

\begin{figure}[!t]
\centering
\includegraphics[width=1.0\linewidth]{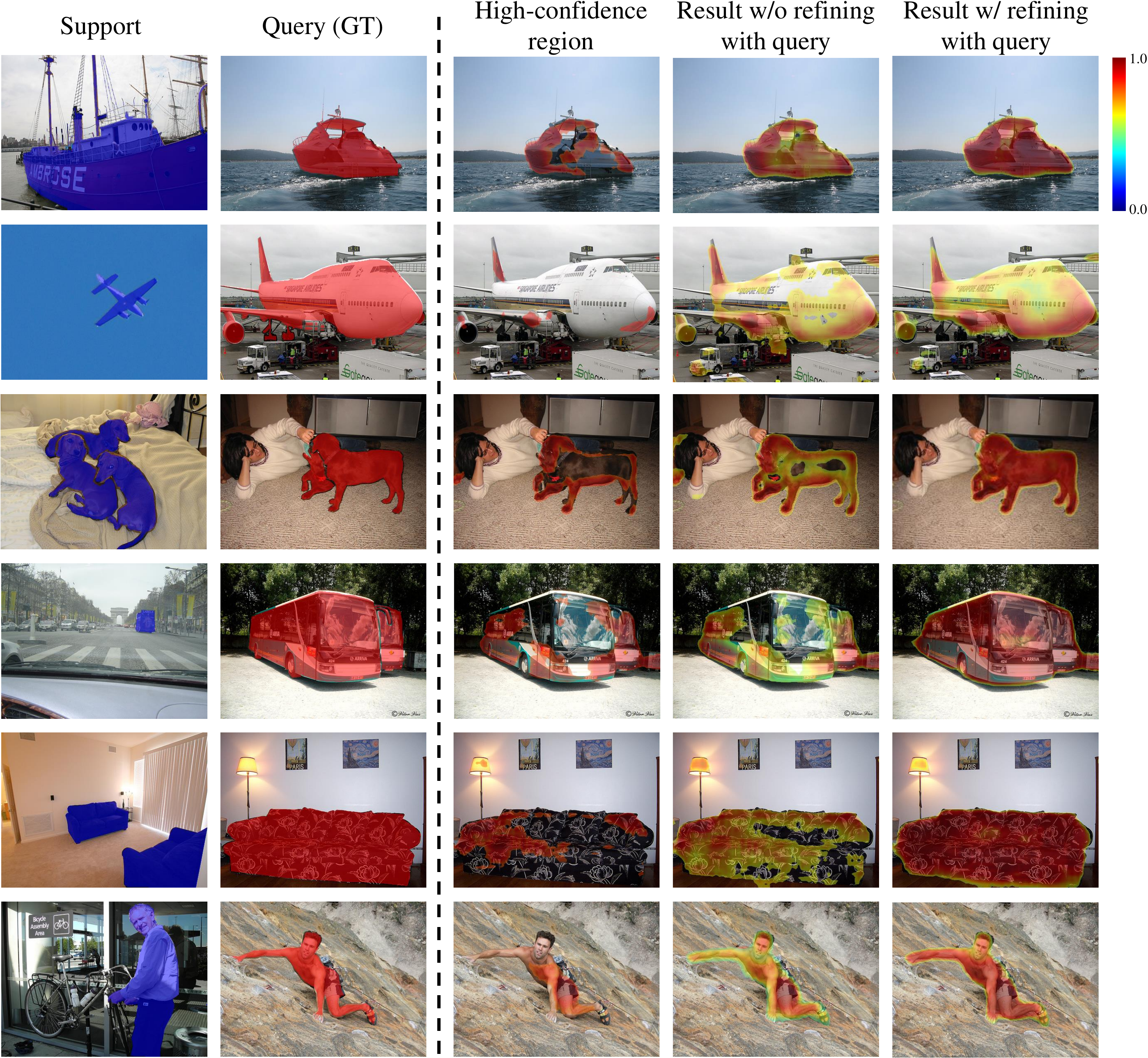}
\caption{Qualitative ablation study of \emph{SROFB}. We visualize the selected high-confidence regions from the query for self-refining based on the prototype-matching paradigm, the predicted masks without and with self-refining using the query, respectively.}
\label{fig:vis_similarity_ablation}
\end{figure}

\subsection{Ablation Study}
We perform extensive ablation studies to investigate the effectiveness of our model. All experiments in this section are performed using ResNet-50 as the backbone in the 1-shot setting on PASCAL-5$^i$.

\smallskip\noindent\textbf{Effectiveness of each functional component.} Table~\ref{table_component} shows the effect of each functional component for few-shot segmentation on novel classes. Our method contains two cores modules, namely \emph{SPFL} and \emph{SROFB}. Using the standard prototype-matching paradigm as the baseline model, we augment the model incrementally by equipping it with single-level \emph{SPFL}, \emph{SPFL} (with hierarchical scheme) and \emph{SROFB}, respectively. As shown in Table~\ref{table_component}, the single-level \emph{SPFL} improves the performance of the baseline model by $2.8\%$, which validates the effectiveness of \emph{SPFL}. Besides, the adopting the hierarchical scheme by \emph{SPFL} can further achieves another $1\%$ performance gain, which reveals the effectiveness of hierarchical pseudo labeling. Compared with the prototype-matching paradigm for segmentation, the proposed \emph{SROFB} yields extra performance gain by 2.1\%, which shows the superiority of \emph{SROFB}. Integrating both \emph{SPFL} and \emph{SROFB}, our method improves the performance of the baseline model from $57.8\%$ to $63.1\%$ on PASCAL-5$^i$, which is a substantial performance gain in few-shot semantic segmentation. 

\begin{table}[!t]
\renewcommand{\arraystretch}{1.2}
    \caption{Ablation study of self-refining of \emph{SROFB} using the query.}
    \label{table_ablation_query}
    \centering
\begin{tabular}{l | c c c c c}
\toprule
 Methods & fold0 & fold1 & fold2 & fold3 & \textbf{Mean} \\ \midrule
\emph{SROFB} (w/o query) & 55.7 & 66.3 & 66.4 & 48.7 & 59.3 \\
\emph{SROFB} (w/ query) & 61.8 & 69.3 & 71.0 & 52.7 & 63.7 \\

\bottomrule
\end{tabular}
\end{table}


    %

\smallskip\noindent\textbf{Effectiveness of self-refining of \emph{SROFB} using the query.} To investigate the effectiveness of self-refining scheme of the proposed \emph{SROFB}, namely refining itself using the high-confidence pixels of the query, we compare the performance of our model between with and without self-refining using the query. The results presented in Table~\ref{table_ablation_query} shows that the self-refining scheme can consistently improve the performance on all splits and achieves $4.4\%$ `Mean' performance gain, which demonstrates the advantage of such design. 

We further perform qualitative evaluation on \emph{SROFB} in Figure~\ref{fig:vis_similarity_ablation}, in which we visualize the selected high-confidence pixels from the query for self-refining based on the prototype-matching paradigm, the predicted masks without and with self-refining using the query, respectively. We can observe that using the prototype-matching paradigm can capture the most similar parts between the support and the query. However, it is challenging for such method to deal with the semantic gap between the support and the query. On the other hand, training \emph{SROFB} based on pure support samples, namely without self-refining on the query, also suffers from the semantic gap between the support and the query. In contrast, Self-refining the \emph{SROFB} leads to much more precise segmentation results.

\begin{figure}[!t]
\centering
\includegraphics[width=1.0\linewidth]{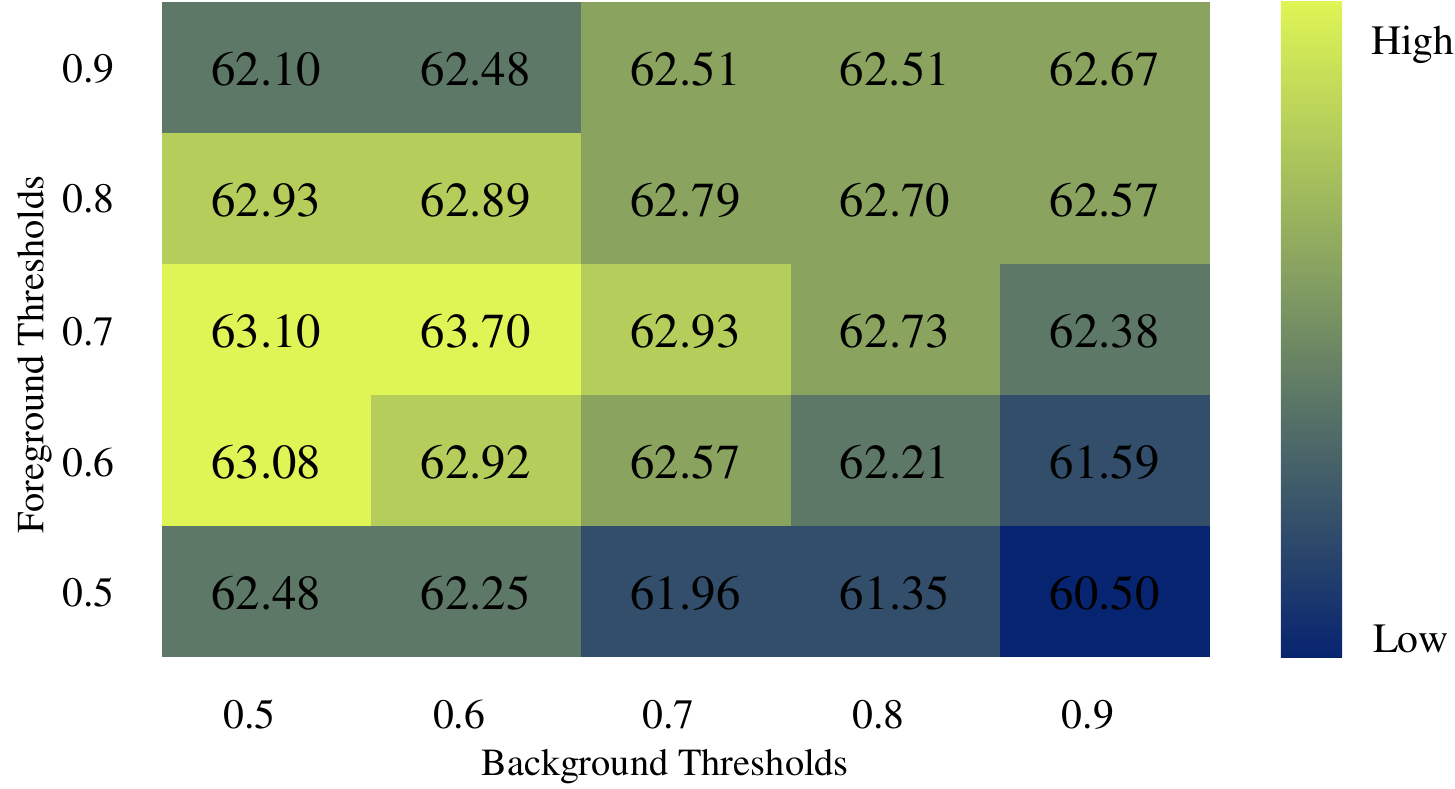}
\caption{Tuning of the selection thresholds for self-refining of \emph{SROFB}.}
\label{fig:query_selection}
\end{figure}

A higher threshold for selecting the high-confidence pixels from the query image results in higher-quality but less training samples for self-refining of \emph{SROFB}. By contrast, a lower threshold leads to sufficient but less reliable training samples. We tune the value of the threshold carefully, as shown in Figure~\ref{fig:query_selection}, based on which we set the foreground threshold and the background threshold as 0.7 and 0.6 in terms of Cosine similarity, respectively.

\begin{figure}[!t]
\centering
\includegraphics[width=0.75\linewidth]{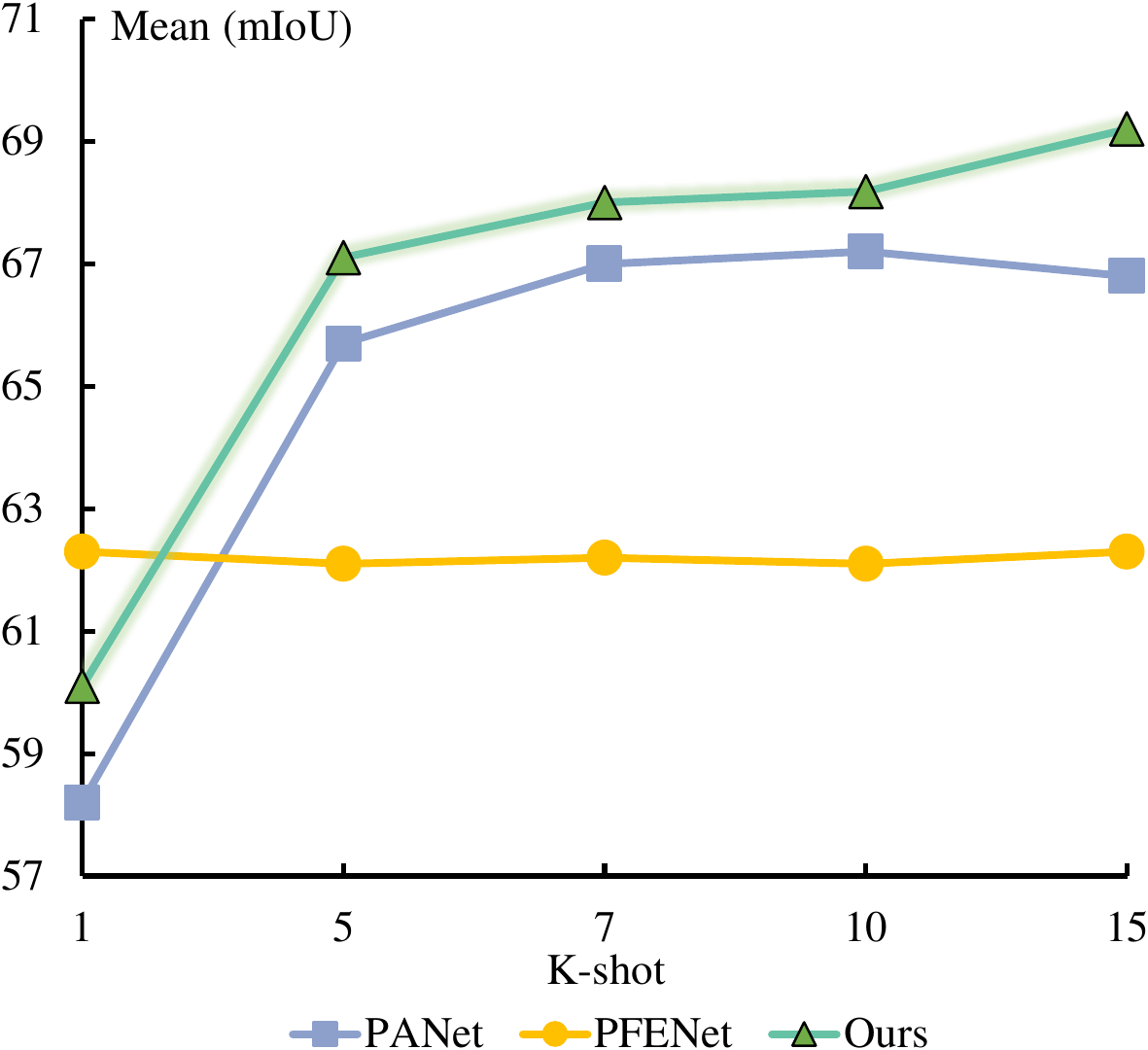}
\caption{Performance of different methods as a function of the number of support images. Compared to an improved version~\cite{yang2021mining} of PANet~\cite{wang2019panet} following the prototype-matching paradigm and PFENet~\cite{tian2020prior} which follows parametric relation-encoding paradigm, our model exhibits better scalability to increasing support samples and can improve the performance steadily. Note that, all methods are trained with 5 shots while performing test using increasing support samples ranging from 1 to 15 samples.}
\label{fig:more_support}
\end{figure}

\smallskip\noindent\textbf{Scalability to increasing support images.\label{sec:ablation}} Most of existing approaches, either following the prototype-matching paradigm like PANet~\cite{wang2019panet} or the parametric relation-decoding paradigm like PFENet~\cite{tian2020prior}, have a potential limitation that they have weaker scalability to increasing support images. They can hardly achieve consistent performance gain as the increase of support samples, as shown in Figure~\ref{fig:more_support}. This is presumably because 1) the inconsistent number of support images between the training and test stages would result in the degradation of the fitting capability of these models~\cite{boudiaf2021few}; 2) both the prototype-matching paradigm and the relation-decoding paradigm are particularly effective in the quite few-shot settings, whilst their performance tends to be saturated much more quickly than our \emph{SROFB}, as shown in Figure~\ref{fig:more_support}. 
In contrast, our \emph{SROFB} does not suffer from such limitation and can improves the performance steadily with the increase of support images.

\smallskip\noindent\textbf{Balance between hierarchical supervision of \emph{SPFL} with multi-granularity pseudo labels.}  
As shown in Equation~\ref{eqn:hierarchy}, we perform hierarchical supervision using the multi-granularity pseudo labels. Different level of supervision is responsible to learn different scale of semantics. We tune the balancing weights ($\lambda_1$, $\lambda_2$ and $\lambda_3$) between three levels of supervision, which is shown in Table~\ref{table_ablation_lambda}.

\begin{table}[!t]
\renewcommand{\arraystretch}{1.2}
    \caption{Parameter tuning of the balancing weights for hierarchical supervision of \emph{SPFL}.}
    \label{table_ablation_lambda}
    \centering
\begin{tabular}{c c c | c c c c c c}
\toprule
 $\gamma_1$ & $\gamma_2$ & $\gamma_3$ & fold0 & fold1 & fold2 & fold3 & \textbf{Mean} \\ \midrule
1.0 & 1.0 & 1.0 & 60.33  & 69.47  & 69.88  & 52.05 & 62.93 \\
1.0 & 0.5 & 0.5 & 60.72  & 69.50  & 69.24  & 52.34  & 62.95 \\
0.5 & 1.0 & 0.5 & 60.60  & 69.51  & 69.73  & 52.41  & 63.06 \\
0.5 & 0.5 & 1.0 & 60.69  & \textbf{69.57}  & 69.93  & 51.96  & 63.04 \\
0.5 & 1.0 & 1.0 & \textbf{61.76}  & 69.31  & \textbf{71.01}  & \textbf{52.71}  & \textbf{63.70} \\

\bottomrule
\end{tabular}
\end{table}

%% file: Conclusion.tex
Existing methods typically treat the novel classes as background during the training on the base classes, which suppresses the feature learning of novel classes. In this paper, we propose to activate the discriminability of novel classes explicitly in both the feature encoding stage and the prediction stage for segmentation, In the feature encoding stage, we design the Semantic-Preserving Feature Learning module (\emph{SPFL}) to first exploit and then retain latent semantics contained in the whole input image, especially those in the background that may belong to novel classes during the training on base classes. To activate the discriminability of novel classes during the prediction stage, we learn the Self-Refined Online Foreground-Background classifier(\emph{SROFB}), which is not only trained with the pixels of support images, but also refined using the high-confidence pixels of the query image. Thus, it is able to adapt to the query image smoothly and bridge the semantic gap between the support and the query. Extensive experiments on PASCAL-5$^i$ and COCO-20$^i$ datasets validated the effectiveness of our proposed approach.